# A Maximal Tractable Class of Soft Constraints

**David Cohen** D.COHEN@RHUL.AC.UK
*Computer Science Department*
*Royal Holloway, University of London, UK*

**Martin Cooper** COOPER@IRIT.FR
*IRIT*
*University of Toulouse III, France*

**Peter Jeavons** PETER.JEAVONS@COMLAB.OX.AC.UK
*OUCL*
*University of Oxford, UK*

**Andrei Krokhin** AK@DCS.WARWICK.AC.UK
*Computer Science Department*
*University of Warwick, UK*

## Abstract

Many researchers in artificial intelligence are beginning to explore the use of soft constraints to express a set of (possibly conflicting) problem requirements. A soft constraint is a function defined on a collection of variables which associates some measure of desirability with each possible combination of values for those variables. However, the crucial question of the computational complexity of finding the optimal solution to a collection of soft constraints has so far received very little attention. In this paper we identify a class of soft binary constraints for which the problem of finding the optimal solution is tractable. In other words, we show that for any given set of such constraints, there exists a polynomial time algorithm to determine the assignment having the best overall combined measure of desirability. This tractable class includes many commonly-occurring soft constraints, such as "as near as possible" or "as soon as possible after", as well as crisp constraints such as "greater than". Finally, we show that this tractable class is maximal, in the sense that adding any other form of soft binary constraint which is not in the class gives rise to a class of problems which is NP-hard.

## 1. Introduction

The constraint satisfaction framework is widely acknowledged as a convenient and efficient way to model and solve a wide variety of problems arising in Artificial Intelligence, including planning (Kautz & Selman, 1992) and scheduling (van Beek, 1992), image processing (Montanari, 1974) and natural language understanding (Allen, 1995).

In the standard framework a *constraint* is usually taken to be a predicate, or relation, specifying the allowed combinations of values for some fixed collection of variables: we will refer to such constraints here as *crisp* constraints. A number of authors have suggested that the usefulness of the constraint satisfaction framework could be greatly enhanced by extending the definition of a constraint to include also *soft* constraints, which allow different measures of desirability to be associated with different combinations of values (Bistarelli et al., 1997, 1999). In this extended framework a constraint can be seen as a *function*, mapping each possible combination of values to a measure of desirability or undesirability. Finding a solution to a set of constraints then means finding an assignment of values to all of the variables which has the best overall combined desirability measure.

**Example 1.1** Consider an optimization problem with $2n$ variables, $v_1, v_2, \ldots, v_{2n}$, where we wish to assign each variable an integer value in the range $1, 2, \ldots, n$, subject to the following restrictions:

- Each variable $v_i$ should be assigned a value that is as close as possible to $i/2$.
- Each pair of variables $v_i$, $v_{2i}$ should be assigned a pair of values that are as close as possible to each other.

To model this situation we might impose the following soft constraints:

- A unary constraint on each $v_i$ specified by a function $\psi_i$, where $\psi_i(x) = (x - i/2)^2$.
- A binary constraint on each pair $v_i, v_{2i}$ specified by a function $\delta_r$, where $\delta_r(x, y) = |x - y|^r$ for some $r \geq 1$.

We would then seek an assignment to all of the variables which minimizes the sum of all of these constraint functions,

$$\sum_{i=1}^{2n} \psi_i(v_i) + \sum_{i=1}^{n} \delta_r(v_i, v_{2i}).$$

□

The cost of allowing additional flexibility in the specification of constraints, in order to model requirements of this kind, is generally an increase in computational difficulty. In the case of crisp constraints there has been considerable progress in identifying classes of constraints which are *tractable*, in the sense that there exists a polynomial time algorithm to determine whether or not any collection of constraints from such a class can be simultaneously satisfied (Bulatov, 2003; Feder & Vardi, 1998; Jeavons et al., 1997). In the case of soft constraints there has been a detailed investigation of the tractable cases for Boolean problems (where each variable has just 2 possible values) (Creignou et al., 2001), but very little investigation of the tractable cases over larger finite domains, even though there are many significant results in the literature on combinatorial optimization which are clearly relevant to this question (Nemhauser & Wolsey, 1988).

The only previous work we have been able to find on the complexity of non-Boolean soft constraints is a paper by Khatib et al. (2001), which describes a family of tractable soft temporal constraints. However, the framework for soft constraints used by Khatib et al. (2001) is different from the one we use here, and the results are not directly comparable. We discuss the relationship between this earlier work and ours more fully in Section 5.

In this paper we make use of the idea of a *submodular function* (Nemhauser & Wolsey, 1988) to identify a general class of soft constraints for which there exists a polynomial time solution algorithm. Submodular functions are widely used in economics and operational research (Fujishige, 1991; Nemhauser & Wolsey, 1988; Topkis, 1998), and the notion of submodularity provides a kind of discrete analogue of convexity (Lovász, 1983).

Submodular functions are usually defined (Nemhauser & Wolsey, 1988) as real-valued functions on sets (which may be viewed as Boolean tuples), but we consider here the more general case of functions on tuples over an arbitrary finite domain (as in Topkis, 1978). We also allow our functions to take infinite values. By establishing a new decomposition result for binary submodular functions of this kind, we obtain a cubic time algorithm to find the optimal assignment for any set of soft constraints which can be defined using them (such as the constraints in Example 1.1). Because our algorithm is specially devised for submodular functions that are expressed as a combination of binary functions, it is much more efficient in this case than existing general algorithms for submodular function minimization (Schrijver, 2000; Iwata et al., 2001).

We give a number of examples to illustrate the many different forms of soft constraint that can be defined using binary submodular functions, and we also show that this class is *maximal*, in the sense that no other form of binary constraint can be added to the class without sacrificing tractability.

## 2. Definitions

To identify a tractable class of soft constraints we will need to restrict the set of functions that are used to specify constraints. Such a restricted set of possible functions will be called a soft constraint *language*.

**Definition 2.1** *Let $D$ and $E$ be fixed sets. A* soft constraint language over $D$ with evaluations in $E$ *is defined to be a set of functions, $\Gamma$, such that each $\phi \in \Gamma$ is a function from $D^k$ to $E$, for some $k \in \mathbb{N}$, where $k$ is called the arity of $\phi$.*

For any given choice of soft constraint language, $\Gamma$, we define an associated soft constraint satisfaction problem, which we will call sCSP($\Gamma$), as follows.

**Definition 2.2** *Let $\Gamma$ be a soft constraint language over $D$ with evaluations in $E$. An instance $\mathcal{P}$ of* sCSP($\Gamma$) *is a triple $\langle V, D, C\rangle$, where:*

- *$V$ is a finite set of* variables, *which must be assigned values from the set $D$.*
- *$C$ is a set of* soft constraints. *Each $c \in C$ is a pair $\langle \sigma, \phi \rangle$ where: $\sigma$ is a list of variables, of length $|\sigma|$, called the* scope *of c; and $\phi$ is an element of $\Gamma$ of arity $|\sigma|$, called the* evaluation function *of c.*

The evaluation function $\phi$ will be used to specify some measure of desirability or undesirability associated with each possible tuple of values over $\sigma$.

To complete the definition of a soft constraint satisfaction problem we need to define how the evaluations obtained from each evaluation function are combined and compared, in order to define what constitutes an optimal overall solution. Several alternative mathematical approaches to this issue have been suggested in the literature:

- In the semiring based approach (Bistarelli et al., 1997, 1999), the set of possible evaluations, $E$, is assumed to be an algebraic structure equipped with two binary operations, satisfying the axioms of a semiring. One example of such a structure is the real interval $[0, 1]$, equipped with the operations min and max, which corresponds to the conjunctive fuzzy CSP framework (Rosenfeld et al., 1976; Ruttkay, 1994). Another example is the set $\{0, 1, 2, \ldots\} \cup \{\infty\}$, equipped with the operations *max* and *plus*, which corresponds to the weighted CSP framework (Bistarelli et al., 1999).

- In the valued CSP approach (Bistarelli et al., 1999), the set of possible evaluations $E$ is assumed to be a totally ordered algebraic structure with a top and bottom element and a single monotonic binary operation known as *aggregation*. One example of such a structure is the set of multisets over some finite ordered set together with a top element, equipped with the operation of multiset union, which corresponds to the lexicographic CSP framework (Bistarelli et al., 1999).

For our purposes, we require the same properties as the valued CSP approach, with the additional requirement that the aggregation operation has a partial inverse, such that evaluations other than the top element may be "cancelled" when occurring on both sides of an inequality. For simplicity, we shall assume throughout this paper that the set of evaluations $E$ is either the set of non-negative integers together with infinity, or else the set of non-negative real numbers together with infinity[1]. Hence, throughout this paper the bottom element in the evaluation structure is 0, the top element is $\infty$, and for any two evaluations $\rho_1, \rho_2 \in E$, the aggregation of $\rho_1$ and $\rho_2$ is given by $\rho_1 + \rho_2 \in E$. Moreover, when $\rho_1 \geq \rho_2$ we also have $\rho_1 - \rho_2 \in E$. (Note that we set $\infty - \infty = \infty$).

The elements of the set $E$ are used to represent different measure of undesirability, or *penalties*, associated with different combinations of values. This allows us to complete the definition of a soft constraint satisfaction problem with the following simple definition of a solution to an instance.

**Definition 2.3** *For any soft constraint satisfaction problem instance $\mathcal{P} = \langle V, D, C\rangle$, an* assignment *for $\mathcal{P}$ is a mapping $t$ from $V$ to $D$. The* evaluation *of an assignment $t$, denoted $\Phi_{\mathcal{P}}(t)$, is given by the sum (i.e., aggregation) of the evaluations for the restrictions of $t$ onto each constraint scope, that is,*

$$\Phi_{\mathcal{P}}(t) = \sum_{\langle\langle v_1, v_2, \ldots, v_k\rangle, \phi\rangle \in C} \phi(t(v_1), t(v_2), \ldots, t(v_k)).$$

*A* solution *to $\mathcal{P}$ is an assignment with the smallest possible evaluation, and the question is to find a solution.*

**Example 2.4** For any standard constraint satisfaction problem instance $\mathcal{P}$ with crisp constraints, we can define a corresponding soft constraint satisfaction problem instance $\widehat{\mathcal{P}}$ in which the range of the evaluation functions of all the constraints is the set $\{0, \infty\}$. For each crisp constraint $c$ of $\mathcal{P}$, we define a corresponding soft constraint $\widehat{c}$ of $\widehat{\mathcal{P}}$ with the same scope; the evaluation function of $\widehat{c}$ maps each tuple allowed by $c$ to 0, and each tuple disallowed by $c$ to $\infty$.

In this case the evaluation of an assignment $t$ for $\widehat{\mathcal{P}}$ equals the minimal possible evaluation, 0, if and only if $t$ satisfies all of the crisp constraints in $\mathcal{P}$. □

1. Many of our results can be extended to more general evaluation structures, such as the *strictly monotonic* structures described by Cooper (2003), but we will not pursue this idea here.

**Example 2.5** For any standard constraint satisfaction problem instance $\mathcal{P}$ with crisp constraints, we can define a corresponding soft constraint satisfaction problem instance $\mathcal{P}^{\#}$ in which the range of the evaluation functions of all the constraints is the set $\{0, 1\}$. For each crisp constraint $c$ of $\mathcal{P}$, we define a corresponding soft constraint $c^{\#}$ of $\mathcal{P}^{\#}$ with the same scope; the evaluation function of $c^{\#}$ maps each tuple allowed by $c$ to 0, and each tuple disallowed by $c$ to 1.

In this case the evaluation of an assignment $t$ for $\mathcal{P}^{\#}$ equals the number of crisp constraints in $\mathcal{P}$ which are violated by $t$. Hence a solution to $\mathcal{P}^{\#}$ corresponds to an assignment which violates the minimal number of constraints of $\mathcal{P}$, and hence satisfies the maximal number of constraints of $\mathcal{P}$. Finding assignments of this kind is generally referred to as solving the Max-CSP problem (Freuder & Wallace, 1992; Larrosa et al., 1999). □

Note that the problem of finding a solution to a soft constraint satisfaction problem is an NP optimization problem, that is, it lies in the complexity class NPO (see Creignou et al., 2001 for a formal definition of this class). If there exists a polynomial-time algorithm which finds a solution to all instances of sCSP($\Gamma$), then we shall say that sCSP($\Gamma$) is *tractable*. On the other hand, if there is a polynomial-time reduction from some NP-complete problem to sCSP($\Gamma$), then we shall say that sCSP($\Gamma$) is *NP-hard*.

**Example 2.6** Let $\Gamma$ be a soft constraint language over $D$, where $|D| = 2$. In this case sCSP($\Gamma$) is a class of *Boolean* soft constraint satisfaction problems.

If we restrict $\Gamma$ even further, by only allowing functions with range $\{0, \infty\}$, as in Example 2.4, then sCSP($\Gamma$) corresponds precisely to a standard Boolean *crisp* constraint satisfaction problem. Such problems are sometimes known as Generalized Satisfiability problems (Schaefer, 1978). The complexity of sCSP($\Gamma$) for such restricted sets $\Gamma$ has been completely characterised, and it has been shown that there are precisely six tractable cases (Schaefer, 1978; Creignou et al., 2001).

Alternatively, if we restrict $\Gamma$ by only allowing functions with range $\{0, 1\}$, as in Example 2.5, then sCSP($\Gamma$) corresponds precisely to a standard Boolean *maximum satisfiability* problem, in which the aim is to satisfy the maximum number of crisp constraints. Such problems are sometimes known as Max-Sat problems (Creignou et al., 2001). The complexity of sCSP($\Gamma$) for such restricted sets $\Gamma$ has been completely characterised, and it has been shown that there are precisely three tractable cases (see Theorem 7.6 of Creignou et al., 2001).

We note, in particular, that when $\Gamma$ contains just the single binary function $\phi_{XOR}$ defined by

$$\phi_{XOR}(x, y) = \begin{cases} 0 & \text{if } x \neq y \\ 1 & \text{otherwise} \end{cases}$$

then sCSP($\Gamma$) corresponds to the Max-Sat problem for the exclusive-or predicate, which is known to be NP-hard (see Lemma 7.4 of Creignou et al., 2001). □

**Example 2.7** Let $\Gamma$ be a soft constraint language over $D = \{1, 2, \ldots, M\}$, where $M \geq 3$, and assume that $\Gamma$ contains just the set of all unary functions, together with the single binary function $\phi_{EQ}$ defined by

$$\phi_{EQ}(x, y) = \begin{cases} 0 & \text{if } x = y \\ 1 & \text{otherwise.} \end{cases}$$

Even in this very simple case it can be shown that sCSP(Γ) is NP-hard, by reduction from the Minimum 3-terminal Cut problem (Dahlhaus et al., 1994). An instance of this problem consists of an undirected graph $(V, E)$ in which each edge $e \in E$ has an associated weight, together with a set of distinguished vertices, $\{v_1, v_2, v_3\} \subseteq V$, known as *terminals.* The problem is to find a set of edges with the smallest possible total weight whose removal disconnects each possible pair of terminals. Such a set is known as a *minimum 3-terminal cut.*

To obtain the reduction to sCSP(Γ), let $I$ be an instance of Minimum 3-Terminal Cut consisting of the graph $\langle V, E\rangle$ with terminals $\{v_1, v_2, v_3\}$. We construct a corresponding instance $\mathcal{P}_I$ of sCSP(Γ) as follows. The variables of $\mathcal{P}_I$ correspond to the set of vertices $V$. For each edge $\{v_i, v_j\} \in E$, add a binary soft constraint with scope $\langle v_i, v_j\rangle$ and evaluation function $\phi_{EQ}$, as above. Finally, for each terminal $v_i \in \{v_1, v_2, v_3\}$, add a unary constraint on the variable $v_i$ with evaluation function $\psi_i$, defined as follows:

$$\psi_i(x) = \begin{cases} 0 & \text{if } x = i \\ |E| + 1 & \text{otherwise} \end{cases}$$

It is straightforward to check that the number of edges in a minimum 3-terminal cut of $I$ is equal to the evaluation of a solution to $\mathcal{P}_I$. □

The examples above indicate that generalizing the constraint satisfaction framework to include soft constraints does indeed increase the computational complexity, in general. For example, the standard 2-Satisfiability problem is tractable, but the soft constraint satisfaction problem involving only the single binary Boolean function, $\phi_{XOR}$, defined at the end of Example 2.6, is NP-hard. Similarly, the standard constraint satisfaction problem involving only crisp unary constraints and equality constraints is clearly trivial, but the soft constraint satisfaction problem involving only soft unary constraints and a soft version of the equality constraint, specified by the function $\phi_{EQ}$ defined in Example 2.7, is NP-hard.

However, in the next two sections we will show that it is possible to identify a large class of functions for which the corresponding soft constraint satisfaction problem is tractable.

## 3. Generalized Interval Functions

We begin with a rather restricted class of binary functions, with a very special structure.

**Definition 3.1** *Let $D$ be a totally ordered set. A binary function, $\phi : D^2 \to E$ will be called a* generalized interval function *on $D$ if it has the following form:*

$$\phi(x, y) = \begin{cases} 0 & \textit{if } (x < a) \vee (y > b); \\ \rho & \textit{otherwise} \end{cases}$$

*for some $a, b \in D$ and some $\rho \in E$. Such a function will be denoted $\eta^{\rho}_{[a,b]}$.*

We can explain the choice of name for these functions by considering the unary function $\eta^{\rho}_{[a,b]}(x, x)$. This function returns the value $\rho$ if and only if its argument lies in the interval $[a, b]$; outside of this interval it returns the value 0.

We shall write $\Gamma_{GI}$ to denote the set of all generalized interval functions on $D$, where $D = \{1, 2, \ldots, M\}$ with the usual ordering.

$$
\begin{array}{cc}
 & y \\
 & \begin{array}{cccccc} 1 & \cdots & b & b+1 & \cdots & M \end{array} \\
x \quad \begin{array}{c} 1 \\ \vdots \\ a-1 \\ a \\ \vdots \\ M \end{array} &
\left(\begin{array}{cccccc}
0 & \cdots & 0 & 0 & \cdots & 0 \\
\vdots & 0 & \vdots & \vdots & 0 & \vdots \\
0 & \cdots & 0 & 0 & \cdots & 0 \\
\rho & \cdots & \rho & 0 & \cdots & 0 \\
\vdots & \rho & \vdots & \vdots & 0 & \vdots \\
\rho & \cdots & \rho & 0 & \cdots & 0
\end{array}\right)
\end{array}
$$

Figure 1: The table of values for the function $\eta^{\rho}_{[a,b]}$

Note that the table of values for any function $\eta^{\rho}_{[a,b]} \in \Gamma_{GI}$ can be written as an $M \times M$ matrix in which all the entries are 0, except for the rectangular region lying between positions $\langle a, 1\rangle$ and $\langle M.b\rangle$, where the entries have value $\rho$, as illustrated in Figure 1. Hence when $\rho = \infty$, a soft constraint with evaluation function $\eta^{\rho}_{[a,b]}$ is equivalent to a crisp constraint which is a particular form of *connected row-convex* constraint (Deville et al., 1999).

The main result of this section is Corollary 3.6, which states that sCSP$(\Gamma_{GI})$ is tractable. To establish this result we first define a weighted directed graph[2] associated with each instance of sCSP$(\Gamma_{GI})$ (see Figure 2).

**Definition 3.2** *Let* $\mathcal{P} = \langle V, \{1, \ldots, M\}, C\rangle$ *be an instance of* sCSP$(\Gamma_{GI})$. *We define the weighted directed graph* $G_{\mathcal{P}}$ *as follows.*

- *The vertices of* $G_{\mathcal{P}}$ *are as follows:* $\{S, T\} \cup \{v_d \mid v \in V,\ d \in \{0, 1, \ldots, M\}\}$.
- *The edges of* $G_{\mathcal{P}}$ *are defined as follows:*
  - *For each* $v \in V$, *there is an edge from* $S$ *to* $v_M$ *with weight* $\infty$;
  - *For each* $v \in V$, *there is an edge from* $v_0$ *to* $T$ *with weight* $\infty$;
  - *For each* $v \in V$ *and each* $d \in \{1, 2, \ldots, M-2\}$, *there is an edge from* $v_d$ *to* $v_{d+1}$ *with weight* $\infty$;
  - *For each constraint* $\langle\langle v, w\rangle, \eta^{\rho}_{[a,b]}\rangle \in C$, *there is an edge from* $w_b$ *to* $v_{a-1}$ *with weight* $\rho$. *These edges are called "constraint edges".*

2. This construction was inspired by a similar construction for certain Boolean constraints described by Khanna et al. (2000).

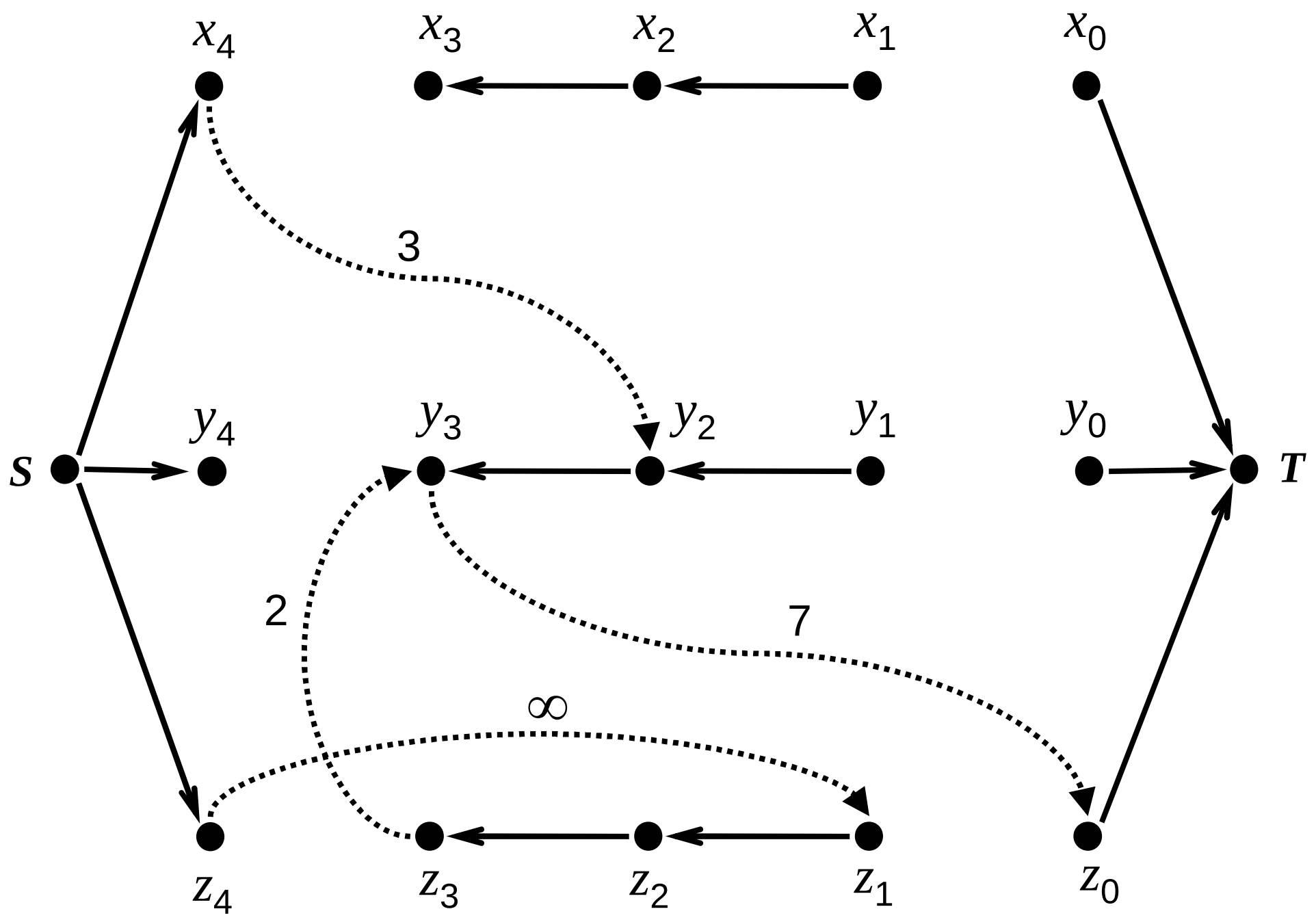

Figure 2: The graph $G_{\mathcal{P}}$ associated with the instance $\mathcal{P}$ defined in Example 3.3. (Note that solid arrows indicate edges with infinite weight.)

**Example 3.3** Let $\mathcal{P} = \langle \{x, y, z\}, \{1, 2, 3, 4\}, C\rangle$ be an instance of sCSP($\Gamma_{GI}$) with the following four constraints:

$$c_1 = \langle\langle y, x\rangle, \eta^{3}_{[3,4]}\rangle$$

$$c_3 = \langle\langle z, y\rangle, \eta^{7}_{[1,3]}\rangle$$

$$c_2 = \langle\langle y, z\rangle, \eta^{2}_{[4,3]}\rangle$$

$$c_4 = \langle\langle z, z\rangle, \eta^{\infty}_{[2,4]}\rangle$$

The corresponding weighted directed graph $G_{\mathcal{P}}$, is shown in Figure 2. □

Any set of edges $C$ in the graph $G_{\mathcal{P}}$ whose removal leaves the vertices $S$ and $T$ disconnected will be called a *cut*. If $C$ is a minimal set of edges with this property, in the sense that removing any edge from $C$ leaves a set of edges which is not a cut, then $C$ will called a *minimal cut*. If every edge in $C$ is a constraint edge, then $C$ will be called a *proper cut*. The *weight* of a cut $C$ is defined to be the sum of the weights of all the edges in $C$.

**Example 3.4** Consider the graph $G_{\mathcal{P}}$ shown in Figure 2. The set $\{\langle y_3, z_0\rangle\}$ is a proper cut in $G_{\mathcal{P}}$ with weight 7, which is minimal in the sense defined above. The set $\{\langle x_4, y_2\rangle, \langle z_3, y_3\rangle\}$ is also a proper cut in $G_{\mathcal{P}}$ with weight 5, which is again minimal in the sense defined above. □

**Proposition 3.5** *Let $\mathcal{P}$ be any instance of* sCSP($\Gamma_{GI}$)*, and let $G_{\mathcal{P}}$ be the associated weighted directed graph, as specified in Definition 3.2.*

1. *For each minimal proper cut in $G_{\mathcal{P}}$ with weight $\Phi$, there is an assignment for $\mathcal{P}$ with evaluation $\Phi$.*

2. *For each assignment $t$ for $\mathcal{P}$ with evaluation $\Phi$, there is a proper cut in $G_{\mathcal{P}}$ with weight $\Phi$.*

**Proof:**

1. Let $C$ be any minimal proper cut of the graph $G_{\mathcal{P}}$, and let $C_S$ be the component of $G_{\mathcal{P}} \setminus C$ connected to $S$. Since $C$ is proper, $C_S$ always contains $v_M$, and never contains $v_0$, so we can define the assignment $t_C$ as follows:

   $$t_C(v) = \min\{d \mid v_d \in C_S\}$$

   By the construction of $G_{\mathcal{P}}$, it follows that:

   $$t_C(v) > d \quad \Leftrightarrow \quad v_d \notin C_S \tag{1}$$

   Now consider any constraint $c = \langle\langle v, w\rangle, \eta^{\rho}_{[a,b]}\rangle$ of $\mathcal{P}$, and its associated edge $e$ in $G_{\mathcal{P}}$. By Definition 3.1 and Equation 1, $\eta^{\rho}_{[a,b]}(t_C(v), t_C(w)) = \rho$ if and only if $v_{a-1} \notin C_S$ and $w_b \in C_S$, and hence if and only if $e$ joins a vertex in $C_S$ to a vertex not in $C_S$. Since $C$ is minimal, this happens if and only if $e \in C$. Hence, the total weight of the cut $C$ is equal to the evaluation of $t_C$.

2. Conversely, let $t$ be an assignment to $\mathcal{P}$, and let $K$ be the set of constraints in $\mathcal{P}$ with a non-zero evaluation on $t$.

   Now consider any path from $S$ to $T$ in $G_{\mathcal{P}}$. If we examine, in order, the constraint edges of this path, and assume that each of the corresponding constraints evaluates to 0, then we obtain a sequence of assertions of the following form:

   $$\begin{array}{rcll} (v_{i_0} > M) & \vee & (v_{i_1} < a_1) & \\ (v_{i_1} > b_2) & \vee & (v_{i_2} < a_2) & \text{for some } b_2 \geq a_1 \\ & \vdots & & \\ (v_{i_{k-1}} > b_k) & \vee & (v_{i_k} < a_k) & \text{for some } b_k \geq a_{k-1} \\ (v_{i_k} > b_{k+1}) & \vee & (v_{i_{k+1}} < 1) & \text{for some } b_{k+1} \geq a_k \end{array}$$

   Since the second disjunct of each assertion contradicts the first disjunct of the next, these assertions cannot all hold simultaneously, so one of the corresponding constraints must in fact give a non-zero evaluation on $t$. Hence, every path from $S$ to $T$ includes at least one edge corresponding to a constraint from $K$, and so the edges corresponding to the set $K$ form a cut in $G_{\mathcal{P}}$. Furthermore, by the choice of $K$, the weight of this cut is equal to the evaluation of $t$.

□

Hence, by using a standard efficient algorithm for the Minimum Weighted Cut problem (Goldberg & Tarjan, 1988), we can find an optimal assignment in cubic time, as the next result indicates.

**Corollary 3.6** *The time complexity of* $\text{sCSP}(\Gamma_{GI})$ *is* $O(n^3|D|^3)$*, where* $n$ *is the number of variables.*

**Proof:** Let $\mathcal{P} = \langle V, D, C\rangle$ be any instance of $\text{sCSP}(\Gamma_{GI})$, and let $G_{\mathcal{P}}$ be the corresponding weighted directed graph. If the minimum weight for a cut in $G_{\mathcal{P}}$ is $\omega < \infty$, then it must be a proper cut, so $\mathcal{P}$ has a solution with evaluation $\omega$, by Proposition 3.5. Moreover, if the minimum weight for a cut in $G_{\mathcal{P}}$ is $\infty$, then the evaluation of every assignment for $\mathcal{P}$ is $\infty$.

Hence we have established a linear-time reduction from $\text{sCSP}(\Gamma_{GI})$ to the Minimum Weighted Cut problem.

Since $G_{\mathcal{P}}$ has $v = |V|(|D|+1)+2$ vertices, and the time complexity of Minimum Weighted Cut is $O(v^3)$ (Goldberg & Tarjan, 1988), the result follows. □

## 4. Submodular Functions

In this section we will consider a rather more general and useful class of functions, as described by Topkis (1978).

**Definition 4.1** *Let* $D$ *be a totally ordered set. A function,* $\phi : D^k \to E$ *is called a* submodular function *on* $D$ *if, for all* $\langle a_1, \ldots, a_k\rangle, \langle b_1, \ldots, b_k\rangle \in D^k$*, we have*

$$\begin{aligned}\phi(\min(a_1,b_1),\ldots,\min(a_k,b_k)) \quad + \quad & \phi(\max(a_1,b_1),\ldots,\max(a_k,b_k)) \\ & \leq \ \phi(a_1,\ldots,a_k) \quad + \quad \phi(b_1,\ldots,b_k).\end{aligned}$$

It is easy to check that all unary functions and all generalized interval functions are submodular. It also follows immediately from Definition 4.1 that the sum of any two submodular functions is submodular. This suggests that in some cases it may be possible to express a submodular function as a sum of simpler submodular functions. For example, for any unary function $\psi : D \to E$ we have

$$\psi(x) \equiv \sum_{d \in D} \eta_{[d,d]}^{\psi(d)}(x,x).$$

For binary functions, the definition of submodularity can be expressed in a simplified form, as follows.

**Remark 4.2** *Let* $D$ *be a totally ordered set. A binary function,* $\phi : D^2 \to E$ *is submodular if and only if, for all* $u, v, x, y \in D$*, with* $u \leq x$ *and* $v \leq y$*, we have:*

$$\phi(u,v) + \phi(x,y) \leq \phi(u,y) + \phi(x,v)$$

*Note that when* $u = x$ *or* $v = y$ *this inequality holds trivially, so it is sufficient to check only those cases where* $u < x$ *and* $v < y$*.*

**Example 4.3** Let $D$ be the set $\{1, 2, \ldots, M\}$ with the usual ordering, and consider the binary function $\pi_M$, defined by $\pi_M(x,y) = M^2 - xy$.

For any $u, v, x, y \in D$, with $u < x$ and $v < y$, we have:

$$\begin{aligned}\pi_M(u,v) + \pi_M(x,y) &= 2M^2 - uv - xy \\ &= 2M^2 - uy - xv - (x-u)(y-v) \\ &\leq \pi_M(u,y) + \pi_M(x,v).\end{aligned}$$

Hence, by Remark 4.2, the function $\pi_M$ is submodular. □

A real-valued $m \times n$ matrix $A$ with the property that

$$A_{uv} + A_{xy} \leq A_{uy} + A_{xv}, \qquad \text{for all } 1 \leq u < x \leq m, \ 1 \leq v < y \leq n$$

is known in operational research as a *Monge matrix* (for a survey of the properties of such matrices and their use in optimization, see Burkard et al., 1996). It is clear from Remark 4.2 that the table of values for a real-valued binary submodular function is a Monge matrix, and conversely, every square Monge matrix can be viewed as a table of values for a binary submodular function.

It was shown by Rudolf and Woeginger (1995) that an arbitrary Monge matrix can be decomposed as a sum of simpler matrices. We now obtain a corresponding result for binary submodular functions, by showing that any binary submodular function can be decomposed as a sum of generalized interval functions. (The result we obtain below is slightly more general than the decomposition result for Monge matrices given by Rudolf and Woeginger (1995), because we are allowing submodular functions to take infinite values.) Using this decomposition result, we will show that the set of unary and binary submodular functions is a tractable soft constraint language.

To obtain our decomposition result, we use the following technical lemma.

**Lemma 4.4** *Let $D$ be a totally ordered set and let $\phi : D^2 \to E$ be a binary submodular function. For any $a, b, c \in D$ such that $a \leq b \leq c$, if there exists $e \in D$ with $\phi(e, b) = 0$, then for all $x \in D$ we have $\phi(x, b) \leq \max(\phi(x, a), \phi(x, c))$.*

**Proof:** Assume that $\phi(e, b) = 0$.

- If $x > e$ then, by the submodularity of $\phi$, we have $\phi(x,b) \leq \phi(x,b) + \phi(e,a) \leq \phi(x,a) + \phi(e,b) = \phi(x,a)$
- If $x < e$ then, by the submodularity of $\phi$, we have $\phi(x,b) \leq \phi(x,b) + \phi(e,c) \leq \phi(e,b) + \phi(x,c) = \phi(x,c)$.
- If $e = x$ then $\phi(x,b) = 0$.

Hence, in all cases the result holds. □

**Lemma 4.5** *Let $D$ be a totally ordered finite set. A binary function, $\phi : D^2 \to E$ is submodular if and only if it can be expressed as a sum of generalized interval functions on $D$. Furthermore, a decomposition of this form can be obtained in $O(|D|^3)$ time.*

**Proof:** By the observations already made, any function $\phi$ which is equal to a sum of generalized interval functions is clearly submodular.

To establish the converse, we use induction on the *tightness* of $\phi$, denoted $\tau(\phi)$, that is, the number of pairs for which the value of $\phi$ is non-zero.

Assume that $\phi$ is a binary submodular function. If $\tau(\phi) = 0$, then $\phi$ is identically zero, so the result holds trivially. Otherwise, by induction, we shall assume that the result holds for all binary submodular functions that have a lower tightness.

To simplify the notation, we shall assume that $D = \{1, 2, \ldots, M\}$, with the usual ordering.

We will say that a value $a \in D$ is *inconsistent* if, for all $y \in D$, $\phi(a,y) = \infty$. If every $a \in D$ is inconsistent, then all values of $\phi$ are $\infty$, so it is equal to the generalized interval function $\eta^{\infty}_{[1,M]}$, and the result holds. Otherwise, if there exists at least one inconsistent value, then we can find a pair of values $a, b \in D$, with $|a-b| = 1$, such that $a$ is inconsistent and $b$ is not inconsistent.

Now define the function $\phi'$ as follows:

$$\phi'(x,y) = \begin{cases} \phi(x,y) & \text{if } x \neq a \\ \phi(b,y) & \text{if } x = a \end{cases}$$

It is straightforward to check that $\phi'$ is submodular and $\phi(x,y) = \phi'(x,y) + \eta^{\infty}_{[a,a]}(x,x)$. Since $\tau(\phi') \leq \tau(\phi)$, it now suffices to show that the result holds for $\phi'$.

By repeating this procedure we may assume that $\phi$ has no inconsistent values, and by symmetry, that the reversed function $\phi^T$, defined by $\phi^T(x,y) = \phi(y,x)$, also has no inconsistent values.

We will say that a value $a \in D$ is *penalized* if, for all $y \in D$, $\phi(a,y) > 0$. If $a$ is penalized, then we set $\mu_a = \min\{\phi(a,y) | y \in D\}$. If $\mu_a = \infty$, then $a$ is inconsistent, so we may assume that $\mu_a < \infty$, and define a new function $\phi'$ as follows:

$$\phi'(x,y) = \begin{cases} \phi(x,y) & \text{if } x \neq a \\ \phi(x,y) - \mu_a & \text{if } x = a. \end{cases}$$

Again it is straightforward to check that $\phi'$ is submodular and $\phi(x,y) = \phi'(x,y) + \eta^{\mu_a}_{[a,a]}(x,x)$. Since $\tau(\phi') \leq \tau(\phi)$, it now suffices to show that the result holds for $\phi'$.

By repeating this procedure we may assume that neither $\phi$ nor $\phi^T$ has any inconsistent or penalized values.

Now if, for all $a, b \in D$, we have $\phi(a,M) = \phi(M,b) = 0$, then, by submodularity, for all $a, b, \in D$, $\phi(a,b) = \phi(a,b) + \phi(M,M) \leq \phi(a,M) + \phi(M,b) = 0$, so $\phi$ is identically 0, and the result holds trivially. Otherwise, by symmetry, we can choose $a$ to be the largest value in $D$ such that $\phi(a,M) \neq 0$. Since $a$ is not penalized, we can then choose $r$ to be the largest value in $D$ such that $\phi(a,r) = 0$. By the choice of $a$, we know that $r < M$, and so we can define $b = r+1$. This situation is illustrated in Figure 3.

For any $x, y \in D$ such that $x \leq a$ and $y \geq b$, we have:

$$\begin{array}{rcll} \phi(x,y) & = & \phi(x,y) + \phi(a,r) & (\phi(a,r) = 0) \\ & \geq & \phi(x,r) + \phi(a,y) & \text{(submodularity)} \\ & = & \phi(x,r) + \max(\phi(a,y), \phi(a,r)) & (\phi(a,r) = 0) \\ & \geq & \phi(x,r) + \phi(a,b) & \text{(Lemma 4.4)} \\ & \geq & \phi(a,b) & \end{array}$$

Hence we can now define a function $\phi'$ as follows:

$$\phi'(x,y) = \begin{cases} \phi(x,y) & \text{if } x > a \vee y < b \\ 0 & \text{if } x = a \wedge y = b \\ \phi(x,y) - \phi(a,b) & \text{otherwise.} \end{cases}$$

It is straightforward to check that $\phi(x,y) = \phi'(x,y) + \eta^{\phi(a,b)}_{[b,a]}(y,x)$. Since $\tau(\phi') < \tau(\phi)$, it only remains to show that $\phi'$ is submodular, and then the result follows by induction. In

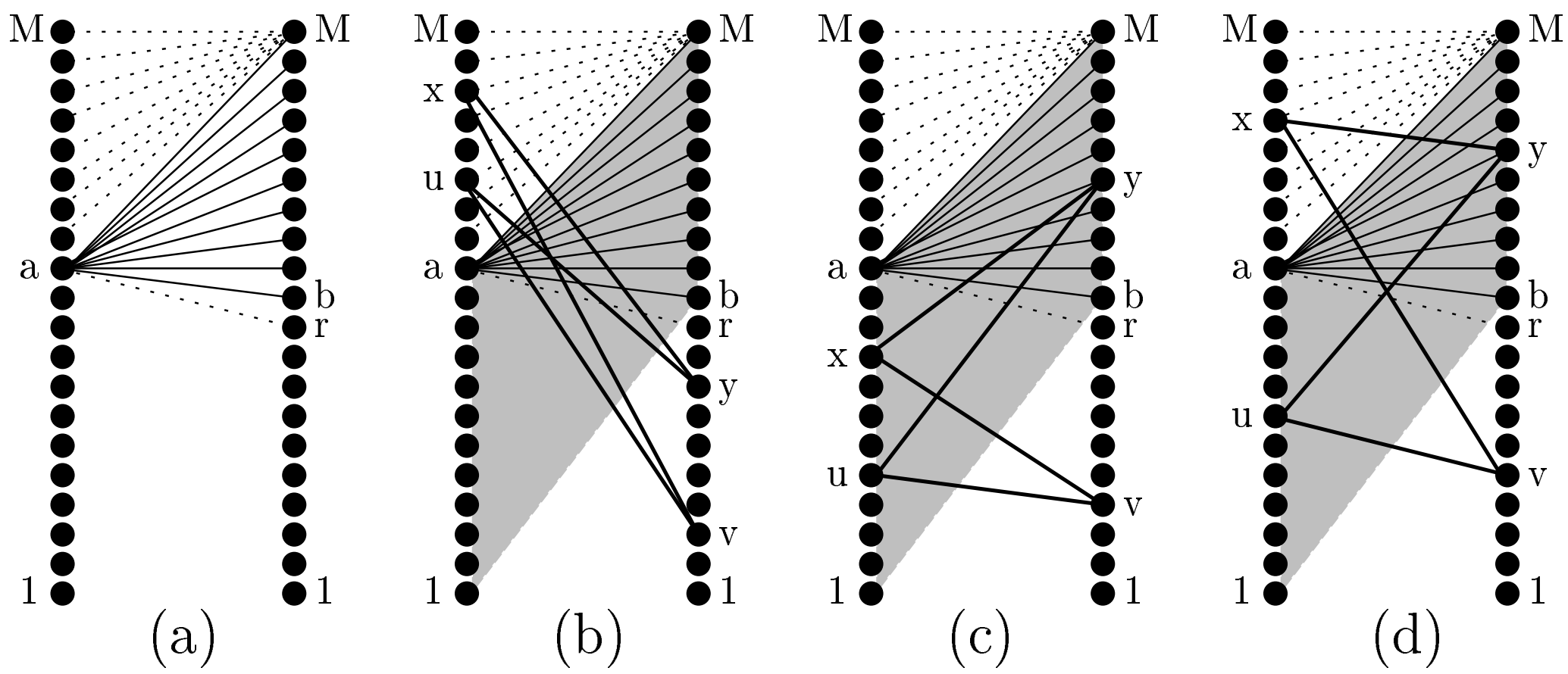

Figure 3: (a) The choice of $a$ and $b$ in the proof of Theorem 4.5. Dotted lines represent known 0 values of $\phi$. Solid lines represent values of $\phi$ known not to be 0. (b-d) Representations of the three cases for the choice of $u, v, x, y$. The filled area represents the non-zero values of the generalized interval constraint subtracted from $\phi$ to obtain $\phi'$.

other words, it suffices to show that for any $u, v, x, y \in D$ such that $u < x$ and $v < y$, we have:

$$\phi'(u,v) + \phi'(x,y) \leq \phi'(u,y) + \phi'(x,v) \tag{2}$$

Replacing $x$ with $u$ in the inequality derived above, we have that whenever $u \leq a$ and $y \geq b$,

$$\phi(u,y) \geq \phi(u,r) + \phi(a,b). \tag{3}$$

The proof of inequality (2) may be divided into four cases, depending on the values of $\phi(a,b)$ and the choice of $u, v, x, y$:

1. $\phi(a,b) = \infty$

   In this case, $\phi'$ differs from $\phi$ only on the pair $\langle a,b \rangle$ (because $\infty - \infty = \infty$). Since $\phi$ is submodular, inequality (2) can only fail to hold if either $\langle x,v \rangle$ or $\langle u,y \rangle$ equals $\langle a,b \rangle$.

   If $\langle x,v \rangle = \langle a,b \rangle$, then, using inequality (3), we know that $\phi(u,y) = \infty$, so $\phi'(u,y) = \infty - \infty = \infty$, and inequality (2) holds.

   If $\langle u,y \rangle = \langle a,b \rangle$ then we have, for all $x > u$ and $y > v$,

$$\begin{array}{rcll} \phi'(u,v) + \phi'(x,y) & = & \phi(u,v) + \phi(x,y) & \\ & \leq & \phi(u,v) + \max(\phi(x,r), \phi(x,M)) & \text{(by Lemma 4.4)} \\ & = & \phi(u,v) + \phi(x,r) & (x > a \Rightarrow \phi(x,M) = 0) \\ & \leq & \phi(u,r) + \phi(x,v) & \text{(by submodularity)} \\ & = & \phi(x,v) & (\text{since } \phi(u,r) = 0) \\ & = & \phi'(x,v) & \\ & \leq & \phi'(u,y) + \phi'(x,v) & \end{array}$$

   so inequality (2) holds.

2. $a < u < x$ or $v < y < b$; (see Figure 3 part (b))

   In this situation we know that inequality (2) holds because $\phi$ and $\phi'$ are identical for these arguments.

3. $u < x \leq a$ or $b \leq v < y$; (see Figure 3 part (c))

   If $u < x \leq a$, then we have:

$$\begin{aligned} \phi'(u,v) &= \phi(u,v) - \rho \\ \phi'(x,v) &= \phi(x,v) - \rho \\ \phi'(u,y) &= \phi(u,y) - \rho' \\ \phi'(x,y) &= \phi(x,y) - \rho' \end{aligned}$$

   where $\rho$ and $\rho'$ are either 0 or $\phi(a,b)$, depending on whether $v$ or $y$ are less than $b$. Inequality (2) follows trivially by cancelling $\rho$ or $\rho'$ or both.

   An exactly similar argument holds if $b \leq v < y$.

4. $u \leq a < x$ and $v < b \leq y$; (see Figure 3 part (d))

   If $u < a$, than by inequality (3) we have $\phi(u,y) - \phi(a,b) \geq \phi(u,r)$, so $\phi'(u,y) \geq \phi(u,r)$. Moreover, if $u = a$, then $\phi(u,r) = 0$, so again $\phi'(u,y) \geq \phi(u,r)$. Hence,

$$\begin{aligned} \phi'(u,v) + \phi'(x,y) &= \phi(u,v) + \phi(x,y) \\ &\leq \phi(u,v) + \max(\phi(x,r), \phi(x,M)) && \text{(by Lemma 4.4)} \\ &= \phi(u,v) + \phi(x,r) && (x > a \Rightarrow \phi(x,M) = 0) \\ &\leq \phi(u,r) + \phi(x,v) && \text{(by submodularity)} \\ &\leq \phi'(u,y) + \phi(x,v) && (\text{since } \phi'(u,y) \geq \phi(u,r)) \\ &\leq \phi'(u,y) + \phi'(x,v) \end{aligned}$$

   so again inequality (2) holds.

Hence, in all cases inequality (2) holds, so $\phi'$ is submodular, and the result follows by induction.

The number of generalized interval functions in the decomposition of a binary submodular function can grow quadratically with $|D|$ (see Example 4.6 below) and the cost of subtracting one binary submodular function from another is also quadratic in $|D|$. Hence a naive algorithm to obtain such a decomposition by calculating the required generalized interval functions and subtracting off each one in turn from the original function will take $O(|D|^4)$ time. However, by taking advantage of the simple structure of generalized interval functions, it is possible to obtain a suitable decomposition in $O(|D|^3)$ time; a possible algorithm is given in Figure 4. The correctness of this algorithm follows directly from the proof of the decomposition result given above. □

**Example 4.6** Consider the binary function $\pi_M$ on $D = \{1, 2, \ldots, M\}$, defined in Example 4.3. When $M = 3$, the values of $\pi_3$ are given by the following table:

| $\pi_3$ | 1 | 2 | 3 |
|---|---|---|---|
| 1 | 8 | 7 | 6 |
| 2 | 7 | 5 | 3 |
| 3 | 6 | 3 | 0 |

**Input:** A binary submodular function $\phi$ on the set {1,2,...,M}
such that neither $\phi$ nor $\phi^T$ has any inconsistent or penalized values

**Output:** A set of generalized interval functions $\{\phi_1, \phi_2, \ldots, \phi_q\}$
such that $\phi(x,y) = \sum_{i=1}^{q} \phi_i(x,y)$

**Algorithm:**

```
for j = 1 to M,  T[j] = 0                            % Initialise list of values to be subtracted
for i = M downto 1                                   % For each row...
    while φ(i,M) > T[M]  do                          % If φ(i,M) not yet zero...
        j = M;  while φ(i,j) > T[j]  do j = j − 1    % Find maximal zero position in row i
        Δ = φ(i,j+1) − T[j+1]                        % Set new value to be subtracted
        output η^Δ_[j+1,i](y,x)                      % Output generalized interval function
        for k = j+1 to M,   T[k] = T[k] + Δ          % Update list of values to be subtracted
    for j = 1 to M,  φ(i,j) = φ(i,j) − T[j]          % Subtract values from this row

for i = 1 to M,  T[j] = 0                            % Initialise list of values to be subtracted
for j = M downto 1                                   % For each column...
    while φ(M,j) > T[M]  do                          % If φ(M,j) not yet zero...
        i = M;  while φ(i,j) > T[i]  do i = i − 1    % Find maximal zero position in column j
        Δ = φ(i+1,j) − T[i+1]                        % Set new value to be subtracted
        output η^Δ_[i+1,j](x,y)                      % Output generalized interval function
        for k = i+1 to M,   T[k] = T[k] + Δ          % Update list of values to be subtracted
    for i = 1 to M,  φ(i,j) = φ(i,j) − T[i]          % Subtract values from this column
```

Figure 4: A decomposition algorithm with time complexity $O(|D|^3)$

Note that:

$$\begin{pmatrix} 8 & 7 & 6 \\ 7 & 5 & 3 \\ 6 & 3 & 0 \end{pmatrix} =$$

$$\begin{pmatrix} 6 & 6 & 6 \\ 0 & 0 & 0 \\ 0 & 0 & 0 \end{pmatrix} + \begin{pmatrix} 0 & 0 & 0 \\ 3 & 3 & 3 \\ 0 & 0 & 0 \end{pmatrix} + \begin{pmatrix} 2 & 0 & 0 \\ 2 & 0 & 0 \\ 2 & 0 & 0 \end{pmatrix} + \begin{pmatrix} 0 & 1 & 0 \\ 0 & 1 & 0 \\ 0 & 1 & 0 \end{pmatrix}$$

$$+ \begin{pmatrix} 0 & 0 & 0 \\ 1 & 1 & 0 \\ 1 & 1 & 0 \end{pmatrix} + \begin{pmatrix} 0 & 0 & 0 \\ 0 & 0 & 0 \\ 1 & 1 & 0 \end{pmatrix} + \begin{pmatrix} 0 & 0 & 0 \\ 1 & 0 & 0 \\ 1 & 0 & 0 \end{pmatrix} + \begin{pmatrix} 0 & 0 & 0 \\ 0 & 0 & 0 \\ 1 & 0 & 0 \end{pmatrix}.$$

Hence,

$$\begin{aligned} \pi_3(x,y) &= \eta^6_{[1,1]}(x,x) + \eta^3_{[2,2]}(x,x) + \eta^2_{[1,1]}(y,y) + \eta^1_{[2,2]}(y,y) \\ &\quad + \eta^1_{[2,2]}(x,y) + \eta^1_{[3,2]}(x,y) + \eta^1_{[2,1]}(x,y) + \eta^1_{[3,1]}(x,y). \end{aligned}$$

In general, for arbitrary values of $M$, we have

$$\pi_M(x,y) = \sum_{d=1}^{M-1} \left( \eta^{M(M-d)}_{[d,d]}(x,x) + \eta^{M-d}_{[d,d]}(y,y) + \sum_{e=1}^{M-1} \eta^1_{[d+1,e]}(x,y) \right)$$

We remark that this decomposition is not unique - other decompositions exist, including the symmetric decomposition $\pi_M(x,y) = \pi'_M(x,y) + \pi'_M(y,x)$, where

$$\pi'_M(x,y) = \sum_{d=1}^{M-1}\left(\eta_{[d,d]}^{\frac{(M^2-d^2)}{2}}(x,x) + \eta_{[d+1,d]}^{\frac{1}{2}}(x,y) + \sum_{e=1}^{d-1}\eta_{[d+1,e]}^{1}(x,y)\right)$$

□

Combining Lemma 4.5 with Corollary 3.6, gives:

**Theorem 4.7** *For any finite soft constraint language $\Gamma$ on a finite totally ordered set $D$, if $\Gamma$ contains only unary or binary submodular functions, then the time complexity of* sCSP($\Gamma$) *is $O(n^3|D|^3)$.*

The next result shows that the tractable class identified in Theorem 4.7 is maximal.

**Theorem 4.8** *Let $\Gamma$ be the set of all binary submodular functions on a totally ordered finite set $D$, with $|D| \geq 2$. For any binary function $\psi \notin \Gamma$,* sCSP($\Gamma \cup \{\psi\}$) *is NP-hard.*

**Proof:** We shall give a reduction from sCSP($\{\phi_{XOR}\}$) to sCSP($\Gamma \cup \{\psi\}$), where $\phi_{XOR}$ is the binary function defined in Example 2.6. It was pointed out in Example 2.6 that sCSP($\{\phi_{XOR}\}$) corresponds to the Max-Sat problem for the exclusive-or predicate, which is known to be NP-hard (Creignou et al., 2001). Hence sCSP($\Gamma \cup \{\psi\}$) is also NP-hard.

To simplify the notation, we shall assume that $D = \{1, 2, \ldots, M\}$, with the usual ordering.

Since $\psi$ is not submodular, there exist $a, b, c, d \in D$ such that $a < b$ and $c < d$ but $\psi(a,c) + \psi(b,d) > \psi(a,d) + \psi(b,c)$.

Choose an arbitrary evaluation $\epsilon$ such that $0 < \epsilon < \infty$, and define $\lambda$ and $\mu$ as follows:

$$\begin{aligned}\lambda &= \min(\psi(a,c), \psi(a,d) + \psi(b,c) + \epsilon)\\ \mu &= \min(\psi(b,d), \psi(a,d) + \psi(b,c) + \epsilon)\end{aligned}$$

It is straightforward to check that

$$\psi(a,d) + \psi(b,c) < \lambda + \mu < \infty. \qquad (4)$$

Now define a binary function $\zeta$ as follows:

$$\zeta(x,y) = \begin{cases} \mu & \text{if } (x,y) = (1,a) \\ \lambda & \text{if } (x,y) = (2,b) \\ \infty & \text{otherwise} \end{cases}$$

and a binary function $\phi$ as follows:

$$\phi(x,y) = \begin{cases} 0 & \text{if } (x,y) = (c,1) \\ \psi(a,d)+1 & \text{if } (x,y) = (c,2) \\ \psi(b,c)+1 & \text{if } (x,y) = (d,1) \\ 0 & \text{if } (x,y) = (d,2) \\ \infty & \text{otherwise} \end{cases}$$

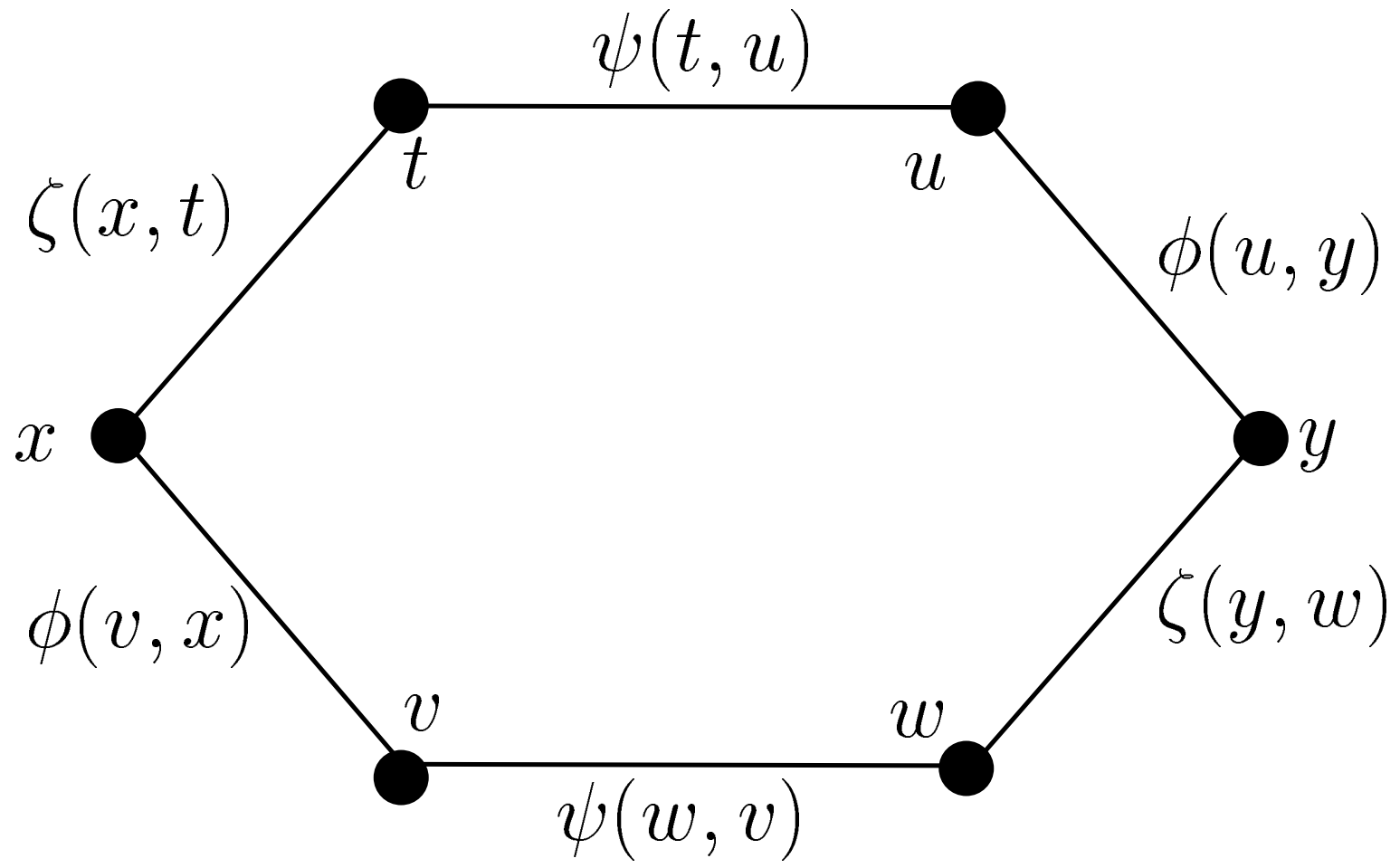

Figure 5: An instance of $\mathrm{sCSP}(\Gamma \cup \{\psi\})$ used to construct a specific soft constraint between variables $x$ and $y$.

It is straightforward to check that both $\zeta$ and $\phi$ are submodular.

Now consider the instance $\mathcal{P}_0$ of $\mathrm{sCSP}(\Gamma \cup \{\psi\})$ illustrated in Figure 5. It is simple but tedious to verify that the combined effect of the six soft constraints shown in Figure 5 on the variables $x$ and $y$ is equivalent to imposing a soft constraint on these variables with evaluation function $\chi$, defined as follows:

$$\chi(x,y) = \begin{cases} \lambda + \mu + \lambda + \mu & \text{if } x, y \in \{1,2\} \text{ and } x = y \\ \lambda + \mu + \psi(a,d) + \psi(b,c) & \text{if } x, y \in \{1,2\} \text{ and } x \neq y \\ \infty & \text{otherwise} \end{cases}$$

Note that, by inequality (4), we have $\lambda + \mu + \psi(a,d) + \psi(b,c) < \lambda + \mu + \lambda + \mu < \infty$.

Now let $\mathcal{P}$ be any instance of $\mathrm{sCSP}(\{\phi_{XOR}\})$. If we replace each constraint $\langle\langle x,y\rangle, \phi_{XOR}\rangle$ in $\mathcal{P}$ with the set of constraints shown in Figure 5 (introducing fresh variables $t,u,v,w$ each time) then we obtain an instance $\mathcal{P}'$ of $\mathrm{sCSP}(\Gamma \cup \{\psi\})$. It is straightforward to check that $\mathcal{P}'$ has a solution involving only the values 1 and 2, and that such solutions correspond exactly to the solutions of $\mathcal{P}$, so this construction gives a polynomial-time reduction from $\mathrm{sCSP}(\{\phi_{XOR}\})$ to $\mathrm{sCSP}(\Gamma \cup \{\psi\})$, as required. □

## 5. Applications

In this section we give a number of examples to illustrate the wide range of soft constraints which can be shown to be tractable using the results obtained in the previous sections.

First we define a standard way to associate a function with a given relation.

**Definition 5.1** *For any $k$-ary relation $R$ on a set $D$, we define an* associated function, *$\phi_R : D^k \to E$, as follows:*

$$\phi_R(x_1, x_2, \ldots, x_k) = \begin{cases} 0 & \textit{if } \langle x_1, x_2, \ldots, x_k \rangle \in R \\ \infty & \textit{otherwise.} \end{cases}$$

By Theorem 4.7, any collection of crisp constraints, where each constraint is specified by a relation $R$ for which $\phi_R$ is unary or binary submodular, can be solved in cubic time, even when combined with other soft constraints that are also unary or binary submodular.

**Example 5.2** The constraint programming language CHIP incorporates a number of constraint solving techniques for arithmetic and other constraints. In particular, it provides a constraint solver for a restricted class of crisp constraints over natural numbers, referred to as *basic constraints* (van Hentenryck et al., 1992). These basic constraints are of two kinds, which are referred to as "domain constraints" and "arithmetic constraints". The domain constraints described by van Hentenryck et al. (1992) are unary constraints which restrict the value of a variable to some specified finite subset of the natural numbers. The arithmetic constraints described by van Hentenryck et al. (1992) have one of the following forms:

$$\begin{array}{ll} aX \neq b & aX \leq bY + c \\ aX = bY + c & aX \geq bY + c \end{array}$$

where variables are represented by upper-case letters, and constants by lower case letters, all constants are non-negative real numbers and $a$ is non-zero.

For each of these crisp constraints the associated function given by Definition 5.1 is unary or binary submodular, hence, by Corollary 3.6, any problem involving constraints of this form can be solved in cubic time. Moreover, any other soft constraints with unary or binary submodular evaluation functions can be added to such problems without sacrificing tractability (including the examples below). □

Now assume, for simplicity, that $D = \{1, 2, \ldots, M\}$.

**Example 5.3** Consider the binary linear function $\lambda$ defined by $\lambda(x, y) = ax + by + c$, where $a, b \in \mathbb{R}^+$.

This function is submodular and hence, by Corollary 3.6, any collection of such binary linear soft constraints over the discrete set $D$ can be solved in cubic time. □

**Example 5.4** The Euclidean length function $\sqrt{x^2 + y^2}$ is submodular, and can be used to express the constraint that a 2-dimensional point $\langle x, y \rangle$ is "as close to the origin as possible". □

**Example 5.5** The following functions are all submodular:

- $\delta_r(x, y) = |x - y|^r$, where $r \in \mathbb{R}$, $r \geq 1$.

  The function $\delta_r$ can be used to express the constraint that: "The values assigned to the variables $x$ and $y$ should be as similar as possible".

- $\delta_r^+(x, y) = (\max(x - y, 0))^r$, where $r \in \mathbb{R}$, $r \geq 1$.

  The function $\delta_r^+$ can be used to express the constraint that: "The value of $x$ is either less than or as near as possible to $y$".

- $\delta_r^{\geq}(x,y) = \begin{cases} |x-y|^r & \text{if } x \geq y \\ \infty & \text{otherwise} \end{cases}$ where $r \in \mathbb{R}$, $r \geq 1$.

  The function $\delta_r^{\geq}$ can be used to express the temporal constraint that: "$x$ occurs as soon as possible after $y$".

□

**Example 5.6** Reconsider the optimization problem defined in Example 1.1. Since $\psi_i$ is unary, and $\delta_r$ is binary submodular (Example 5.5), this problem can be solved in cubic time, using the methods developed in this paper.

Let $\mathcal{P}$ be the instance with $n = 3$ and $r = 2$. The values of $\delta_2$ are given by the following table:

| $\delta_2$ | 1 | 2 | 3 |
|---|---|---|---|
| 1 | 0 | 1 | 4 |
| 2 | 1 | 0 | 1 |
| 3 | 4 | 1 | 0 |

Hence,

$$\begin{aligned} \delta_2(x,y) &= \eta^1_{[3,2]}(y,x) + \eta^1_{[2,1]}(y,x) + \eta^2_{[3,1]}(y,x) \\ &\quad + \eta^1_{[3,2]}(x,y) + \eta^1_{[2,1]}(x,y) + \eta^2_{[3,1]}(x,y) \end{aligned}$$

Using this decomposition for $\delta_2$, we can construct the graph $G_{\mathcal{P}}$ corresponding to the instance $\mathcal{P}$, as shown in Figure 6.

The minimum weight of any cut in this graph is $\frac{11}{4}$, and hence the optimal evaluation of any assignment for $\mathcal{P}$ is $\frac{11}{4}$.

One of the several possible cuts with this weight is indicated by the gray line across the graph, which corresponds to the solution $v_1 = 1$, $v_2 = 1$, $v_3 = 2$, $v_4 = 2$, $v_5 = 3$, $v_6 = 3$. □

Note that some of the submodular functions defined in this section may appear to be similar to the soft simple temporal constraints with semi-convex cost functions defined and shown to be tractable by Khatib et al. (2001). However, there are fundamental differences: the constraints described by Khatib et al. (2001) are defined over an infinite set of values, and their tractability depends crucially on the aggregation operation used for the costs being *idempotent* (i.e., the operation *min*). In this paper we are considering soft constraints over finite sets of values, and an aggregation operation which is *strictly monotonic* (e.g., addition of real numbers), so our results cannot be directly compared with those in the paper by Khatib et al. (2001).

## 6. Conclusion

As we have shown with a number of examples, the problem of identifying an optimal assignment for an arbitrary collection of soft constraints is generally NP-hard. However, by making use of the notion of submodularity, we have identified a large and expressive class of soft constraints for which this problem is tractable. In particular, we have shown that *binary* soft constraints with the property of submodularity can be solved in cubic time. By making use of this result, it should be possible to extend the range of optimisation problems that can be effectively solved using constraint programming.

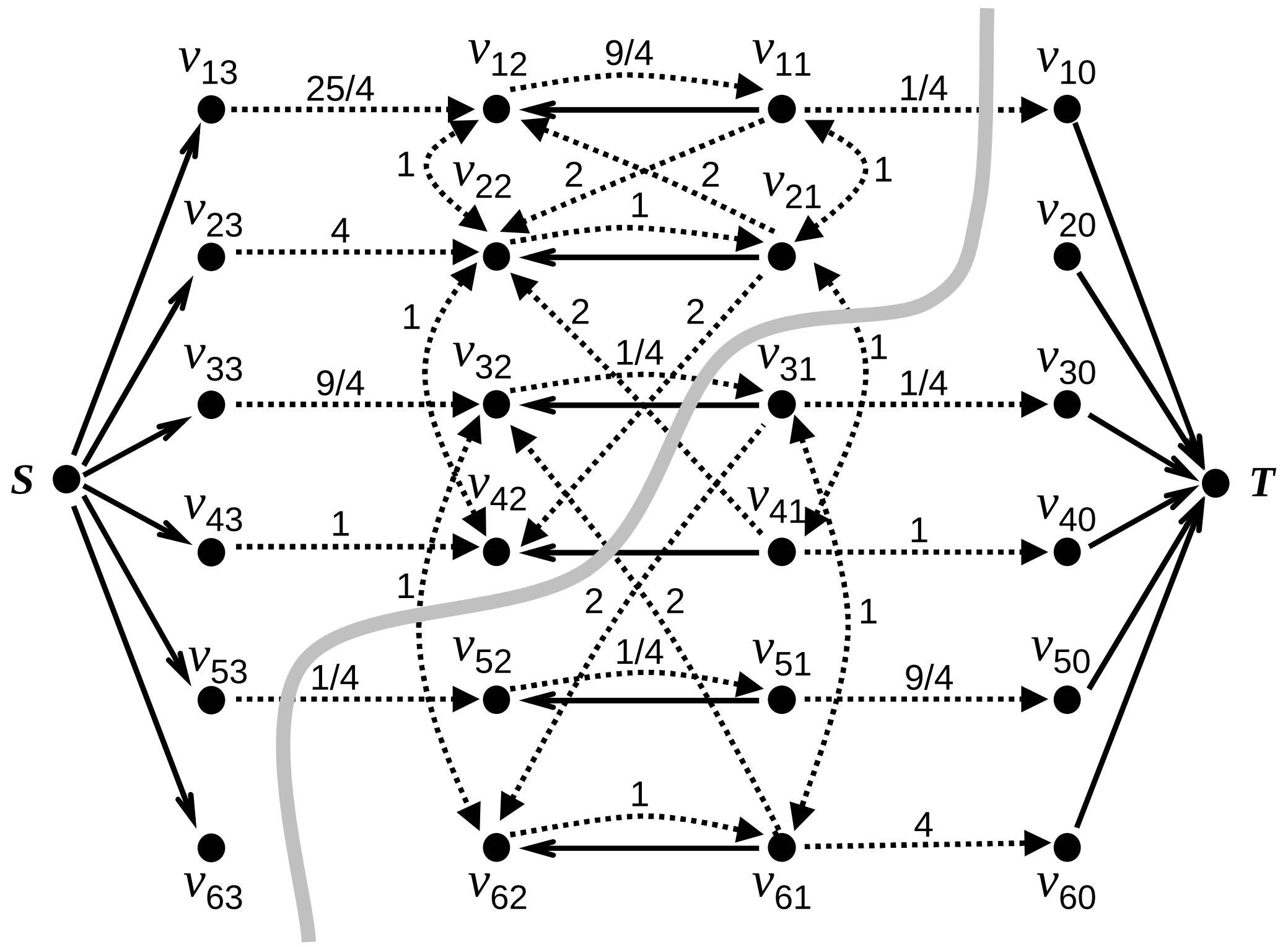

Figure 6: The graph $G_{\mathcal{P}}$ associated with the instance $\mathcal{P}$ defined in Example 5.6.

From a theoretical perspective, this paper gives the first complete characterisation of a tractable class of soft constraints over a finite set of values with more than two elements. We are confident that the methods developed here can be extended to identify other tractable cases, and hence to begin a systematic investigation of the computational complexity of soft constraint satisfaction. A first step in this direction has been taken by Cohen et al. (2003).

We believe that this work illustrates once again the benefit of interaction between research on constraint satisfaction and more traditional research on discrete optimization and mathematical programming: the notion of submodularity comes from mathematical programming, but the idea of modelling problems with binary constraints over arbitrary finite domains comes from constraint programming. By combining these ideas, we obtain a flexible and powerful modelling language with a provably efficient solution strategy.

## 7. Acknowledgments

An earlier version of this paper, omitting many of the proofs, was presented at the International Joint Conference on Artificial Intelligence in 2003. This research was partially supported by the UK EPSRC grant GR/R81213/01.